# TOWARDS HARD AND SOFT SHADOW REMOVAL VIA DUAL-BRANCH SEPARATION NETWORK AND VISION TRANSFORMER

JIAJIA LIANG, BOWEN ZHANG, JINGWEN DENG, PATRICK P. K. CHAN*

Shien-Ming Wu School of Intelligent Engineering, South China University of Technology, China
E-MAIL:202164020270@mail.scut.edu.cn, 202164690534@mail.scut.edu.cn, jingwen_deng@foxmail.com, patrickchan@scut.edu.cn

**Abstract:**
Shadow removal is an important task in computer vision, as shadows in real-world scenes affect image color and brightness, making perception and texture recognition more challenging. Many existing methods may not adequately distinguish between hard and soft shadows, resulting in a lack of targeted processing for each type. To order to tackle this problem, we propose a dual-path model that separates the processing of hard and soft shadows using specifically designed loss functions. The model includes a shadow classification module to guide each shadow type through the appropriate path. By integrating a Vision Transformer with UNet++, our approach improves edge detail and feature fusion. On the ISTD dataset, our model achieves an RMSE of 2.905, outperforming state-of-the-art single-path methods.

## 1. Introduction

Shadow removal has long been a fundamental research topic in computer vision. In real-world scenes, shadows cast by objects can alter the chromaticity and brightness of specific areas in an image. They can also affect texture, posing challenges for chromaticity perception and texture recognition. Consequently, shadow removal is pivotal for tasks such as object detection and recognition, image segmentation, target tracking, and scene understanding .

Recent advancements in deep learning have significantly improved shadow removal techniques [1], particularly for hard shadows. Hard shadows have distinct, sharp edges, making their boundaries easier to identify and mask, while soft shadows have diffused edges that blend into their surroundings, making them less distinct and more challenging to detect. For hard shadows, the primary challenge lies in naturally restoring chromaticity and texture at shadow edges. Since hard shadow regions often exhibit uniform chromaticity and brightness, reliance on local information is typically sufficient for shadow removal tasks.

Current state-of-the-art models [13] have demonstrated impressive performance in shadow removal. However, these methods apply treat both hard and soft shadows with the same techniques and overlook the distinct challenges associated with each. For instance, a Gaussian filter based method performs well on hard shadows but it is less suitable for soft shadows, where blurred edges reduce the filter's effectiveness. Consequently, while ISTD [9] and SRD [14] perform well on datasets dominated by hard shadows, they are less effective on datasets emphasizing soft shadows, such as LRSS [15]. This highlights the need for advanced methods capable of distinguishing and handling both hard and soft shadows to improve shadow removal across diverse scenarios.

To address these challenges, this paper proposes a dual-path mechanism that processes hard and soft shadows separately. Our model's generator is designed with two dedicated paths: one optimized for hard shadows and the other for soft shadows. Each path employs specialized loss functions for independent training, enabling tailored processing for each shadow type. In our approach, a shadow image is first fed into a classifier that determines whether the shadows in the image are hard or soft. The image is then processed through both specialized paths to generate two shadow-free outputs. Finally, based on the classifier's results, these outputs are merged into the final shadow-removed image using predefined fusion parameters. This strategy allows for differentiated treatment of hard and soft shadows, reducing errors associated with single-path methods. Furthermore, it effectively handles images with ambiguous shadow types, ensuring robust performance across various scenarios.

The main contributions of this study are summarized as follows:
1) We propose a novel dual-path mechanism for shadow removal, which processes soft and hard shadows

separately. This approach improves the model's ability to handle diverse shadow types in complex scenes.
2) To the best of our knowledge, this is the first work to integrate Vision Transformer with UNet++ for shadow removal. This combination enhances detail processing, particularly around shadow edges, through a multi-layer feature fusion mechanism.
3) Comprehensive experiments and ablation studies on the ISTD dataset demonstrate that our method outperforms existing state-of-the-art models, validating the efficacy of our proposed approach.

## 2. Related Works

Traditional physical methods for image shadow removal primarily rely on in-depth analysis of lighting models, shadow colors, and scene geometry, aiming to simulate and eliminate shadow effects in images. However, these methods exhibit clear limitations, including strong dependence on environmental lighting conditions and scene geometry, as well as insufficient adaptability when faced with complex and variable real-world scenes. Moreover, the accuracy and robustness of such approaches are also limited when encountering shadows of different types and depths. In contrast to these traditional methods, our proposed approach is capable of effectively handling various types of shadows, achieving efficient and accurate shadow removal effects for both hard and soft shadows.

In the last few years, there has been notable achievement in deep learning methods for the task of image shadow removal. For example, Le et al. and others proposed a new method combining linear illumination transformations and deep learning, which effectively removes shadows from a single image by predicting shadow parameters and shadow masks, significantly improving the accuracy and quality of shadow removal. Similarly, RYO [9] and others developed a new model named CANet, which integrates shadow detection and shadow removal networks, and introduces a ColorBlock structure and trains through an extended dataset, achieving high-precision shadow removal. Furthermore, methods using Generative Adversarial Networks (GAN) have also been applied to shadow removal, where RIS-GAN [10] framework, by exploring residual and illumination information in the shadow removal process, combines rough shadow-free images, estimated negative residual images, and inverse illumination maps to generate detailed shadow-free images. Cun et al. [11] proposed DHAN and SMGAN as well as ST-CGAN [12] further refined the task of shadow removal. However, despite these models making progress in dealing with different types of shadows, their generalization ability remains limited. In contrast, our model can differentiate based on the softness and hardness of shadows, significantly reducing errors that might be introduced by relying on a single processing strategy.

Compared to our method, DC-ShadowNet [13] adopts a similar approach. They designed a network guided by unsupervised learning and a domain classifier, specifically aimed at removing hard and soft shadows from single images. Unlike traditional techniques, DC-ShadowNet [13] innovates by utilizing novel unsupervised loss functions such as physics-based shadow-free chromaticity, shadow-robust perceptual features, and boundary smoothness loss, combined with a domain classifier to guide the generator and discriminator to focus on shadow areas. However, this method does not significantly distinguish between hard and soft shadows but attempts to handle all types of shadows by incorporating traditional physical shadow removal methods into its loss function. Our method, on the other hand, employs two parallel paths trained for diverse shadow profiles in images. Through this differentiated treatment, our model can more effectively adapt to different types of shadow images, thereby enhancing the performance and applicability of shadow removal.

## 3. Proposed Methods

Our method classifies shadow images into hard and soft shadows, leveraging a shadow removal network consisting of three main components: a classifier ($\Phi_s^h$), a generator, and a discriminator. The generator is designed with two distinct pathways, each comprising a Swin Transformer and a UNet++ with separate parameter configurations. Each UNet++ features an encoder-decoder structure. For a given input shadow image ($I_{shadow}$), the outputs include a shadow-free image ($I_{output}$), a shadow mask ($I_{shadow\_mask}$), shadow-free chromaticity ($I_{shadow}^{chr}$), output chromaticity images ($I_{shadow}^{chr}$), and a shadow edge mask ($I_{shadow\_mask\_edge}$). The architecture of our method is shown in Figure 1.

The classifier ($\Phi_s^h$) produces fusion weight ratios that are used to combine the outputs from two specialized pipelines: one for hard shadows and another for soft shadows. To train the generator and discriminator effectively, we employ tailored loss functions for each pipeline. In the soft shadow pipeline, a shadow-free chromaticity loss ($L_{chr}$) is calculated using physical methods based on the original ($I_{shadow}$) and the shadow-free ($I_{output}$),. For the hard shadow pipeline, a shadow edge loss function is derived from the shadow mask using a Gaussian filter to enhance edge precision.

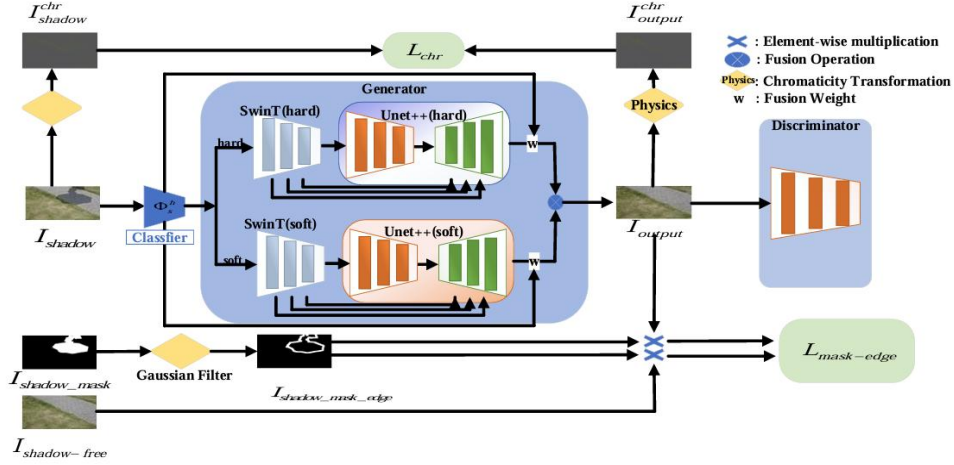

**FIGURE 1.** The architecture of our method

A pre-trained classification network first determines the shadow type for a given image ($I_{shadow}$), classifying it as either a hard or soft shadow. The classifier outputs a binary probability distribution, $[P_{hard}, P_{soft}]$, indicating the likelihood of the shadow type. The image is then propagated simultaneously through two separate pipelines: one designed for hard shadows and the other for soft shadows. Within each pipeline, a Transformer architecture extracts hierarchical feature maps, which are subsequently processed by a UNet++ model. The outputs from both pipelines, representing potential shadow-free images, are merged through a fusion layer. This layer weights the outputs based on the classifier's initial probability distribution $[P_{hard}, P_{soft}]$. The resulting ($I_{output}$) is paired with a real shadow-free image ($I_{shadow\_free}$) and fed into a discriminator. The discriminator's objective is to differentiate between ($I_{output}$) and ($I_{shadow\_free}$).

To address the distinct challenges posed by hard and soft shadows, we incorporated additional loss functions beyond standard ones such as adversarial loss, similarity loss, and perceptual loss. Specifically, two custom loss functions were introduced to enhance the training for each shadow type. For the soft shadow pipeline, a chromaticity loss calculates the difference in color characteristics between the shadow-free chromaticity of the input shadow image and the output shadow-free image. For the hard shadow pipeline, a shadow mask edge loss emphasizes precision at shadow boundaries during the loss computation process.

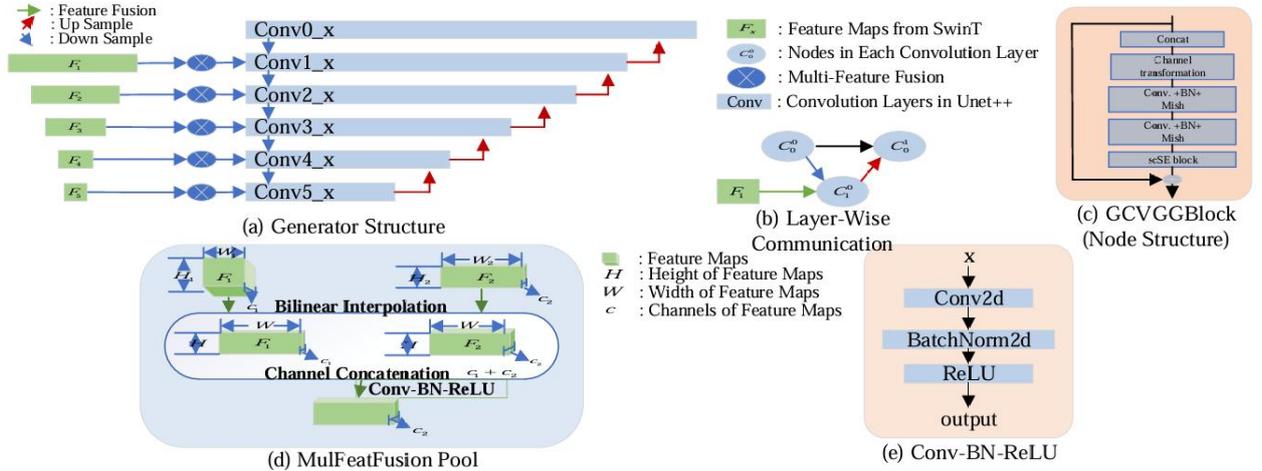

**FIGURE 2.** Overview of the proposed multi-scale fusion network

## 3.1 Multi-Feature Fusion Structure

Figure 2 illustrates the architecture of a multi-scale fusion network, designed to process feature maps extracted from an input source. The network comprises several key components to enhance image analysis. It employs multiple convolutional layers integrated with upsampling and downsampling strategies, facilitating feature fusion across different scales. Layer-wise communication ensures efficient feature map flow, while the GCVGG block processes the detailed feature extraction at each node. The Multi-FeatFusion Pool module merges feature maps using bilinear interpolation and channel concatenation, followed by convolutional operations to reduce channel dimensions. These components collectively leverage multi-scale information for improved performance.

The generator architecture is divided into three primary modules: the feature extractor, the encoder-decoder, and the multi-feature fusion module. The feature extractor leverages the Swin Transformer, which employs hierarchical representations and shifted window-based self-attention mechanisms to extract features across varying scales. This enables the network to capture intricate patterns necessary for spatial analysis and feature integration across multiple levels. After feature extraction, the resulting feature maps are fused at different decoder stages using a multi-feature fusion module, enhancing the model's ability to combine complementary features throughout the network.

A critical component of this architecture is the Multi-FeatFusion Pool module, which merges features from different network levels. It aligns the spatial dimensions of two feature maps through bilinear interpolation before concatenating them along the channel dimension. This concatenated output undergoes further processing via a Conv-BN-ReLU layer that incorporates convolution, batch normalization, and ReLU activation to compress the channel dimensions. The process ensures that the fused features are optimally integrated with the decoder. The convolution operation can be mathematically expressed as:

$$o = \delta(N(Conv(x)))\#(1)$$

where $Conv$ represents Convolution layer, $N$ denotes batch normalization, and $\delta$ is the ReLU activation function. This fusion process plays a pivotal role in improving the model's reconstruction and prediction capabilities.

The encoder-decoder framework, which is the backbone of the generator responsible for shadow removal, is based on the UNet++ architecture. This structure includes five layers of descending convolution nodes, where the number of nodes decreases progressively from six in the initial layer to one in the final layer. Each node comprises a residual block that includes convolutional layers, batch normalization, Mish activation functions, and scSE blocks. The flow of information within the network is carefully orchestrated. The primary node in each layer receives downsampled feature maps from the preceding layer, merges them with incoming data, and upscales the output for subsequent nodes. Intermediate nodes in each layer process local feature maps without additional external inputs, ensuring efficient and focused feature propagation across the network.

## 3.2 Shadow Chromaticity Loss

Soft shadows are significantly influenced by complex lighting conditions, such as the direction, intensity, and color of light. These factors introduce variations in color and brightness within shadowed areas, especially in transition regions between shadowed and non-shadowed areas, where chromaticity changes are more pronounced. To accurately restore these transition regions, we aim to ensure chromaticity consistency between shadowed and non-shadowed images. For this purpose, we incorporate the Shadow Chromaticity Loss into the generator training process for soft shadow paths.

The Shadow Chromaticity Loss is a specialized loss function for image shadow removal and consists of two main steps: entropy minimization and illumination compensation. First, the RGB channel values of the input shadow image are normalized as follows:

$$[X',Y',Z'] = R(X + Y + Z), R \in [X, Y, Z]\#(2)$$

where $X'$, $Y'$, and $Z'$ represent the values of the three RGB channels. The normalized image is then mapped into logarithmic chromaticity space:

$$[\rho_1, \rho_1, \rho_3] = [\log(X'), \log(I'), \log(Z')]\#(3)$$

Next, the two-dimensional log chromaticity image is projected onto a one-dimensional space for various angles between 1° and 180°. The entropy of each projection is calculated, and the projection direction with the lowest entropy is identified as optimal. Using this direction, a shadow-free chromaticity image is generated. However, this process may introduce a color shift due to the projection.

To address the color shift, the brightest set of pixels from the input image is selected, and the chromaticity image is adjusted to reflect the original lighting conditions of the non-shadowed regions. This adjustment ensures the chromaticity image accurately represents the true colors of the scene. Finally, a physically-based shadow-free chromaticity image is used to guide the shadow removal generator through the chromaticity loss function. This ensures that the output image achieves chromaticity consistency with the shadow-free chromaticity image, resulting in more accurate and realistic shadow removal.

## 3.3 Shadow Mask Edge Loss

Hard shadows are characterized by sharp boundaries, which create a distinct contrast between shadowed and non-shadowed areas. To ensure that the texture and color at these edges are accurately restored during shadow removal, we introduce a shadow mask edge loss in the generator training for hard shadow paths. This loss function is specifically designed to focus on the boundary regions between shadowed and non-shadowed areas. To achieve this, we use a combination of Gaussian and Sobel filters to generate a shadow boundary mask that highlights these edges:

$$M_{\text{rete}} = \begin{cases} 1 & G(\text{Edge}(M_y)) > 0 \\ 0 & \text{otherwise} \end{cases} \quad (4)$$

where G stands for Gaussian filter.

Using the shadow boundary mask $M_{edge}$, the difference between the ground truth shadow-free image $I_g$ and the generated shadow-free image $I_{output}$ is calculated. The element-wise product of these differences is then computed, and the Euclidean distance of the resulting product is summed across all pixel positions $i$ to obtain the shadow edge loss function. This approach enables more precise handling of shadow edges, ensuring that the generated images appear natural and free of noticeable shadow artifacts. The shadow edge loss function is defined as:

$$L_{\text{edge}} = \sum_i^N \|(I_{gt} - I_{\text{output}}) \cdot M_{\text{edge}}\| \#(5)$$

By emphasizing the boundaries, this method improves the restoration of sharp transitions, resulting in visually realistic images with well-preserved edge details.

## 4. Experiments

To evaluate the effectiveness of our approach, we conducted experiments on the ISTD dataset [9], the LRSS dataset, and an expanded version of the LRSS dataset. The ISTD dataset contains 1,330 training tuples and 540 testing tuples, with each tuple comprising a shadow image, shadow mask, and shadow-free image. This dataset primarily features hard shadows, accounting for approximately 70%–80% of the images. In contrast, the LRSS dataset, originally introduced by Yeying Jin [23], focuses on soft shadows but is significantly smaller, comprising only 34 image groups, each including shadow images, shadow masks, and shadow-free images. To address the limitations of the LRSS dataset's small size, we expanded it using the MaskGAN [24] methodology to generate additional training data. This expansion aimed to enhance the dataset's diversity and improve the robustness of our model during training.

TABLE 1. Quantitative comparison with the SOTA methods on the ISTD dataset. '-' represents the unavailable results.

| Methods | Citation | RMSE | | | SSIM | | | PSNR | | |
|---|---|---|---|---|---|---|---|---|---|---|
| | | Shadow | No Shadow | Average | Shadow | No Shadow | Average | Shadow | No Shadow | Average |
| Guo et al. [25] | CVPR/2011 | 18.65 | 7.76 | 9.26 | 0.964 | 0.975 | 0.919 | 27.76 | 26.44 | 23.08 |
| Deshadownet [4] | CVPR/2017 | 17.96 | 6.53 | 8.47 | - | - | - | - | - | - |
| ST-CGAN [5] | CVPR/2018 | 10.11 | 5.76 | 6.47 | 0.981 | 0.959 | 0.932 | 33.93 | 30.18 | 27.90 |
| Mask-ShadowGAN [6] | ICCV/2019 | 10.57 | 5.91 | 6.67 | 0.980 | 0.959 | 0.928 | 31.73 | 29.02 | 26.36 |
| DSC [7] | CVPR/2018 | 8.45 | 5.03 | 5.59 | 0.984 | 0.969 | 0.944 | 34.64 | 31.26 | 29.00 |
| AngularGAN [2] | CVPRW/2019 | 9.78 | 7.67 | 8.16 | - | - | - | - | - | - |
| DHAN [16] | AAAI/2020 | 7.49 | 5.30 | 5.66 | 0.988 | 0.971 | 0.954 | 35.53 | 31.05 | 29.11 |
| RIS-GAN [20] | AAAI/2020 | 8.99 | 6.33 | 6.95 | - | - | - | - | - | - |
| CLA-GAN [21] | PG/2020 | 9.01 | 6.25 | 6.62 | - | - | - | - | - | - |
| AEF [19] | CVPR/2021 | 7.98 | 5.54 | 5.94 | 0.974 | 0.880 | 0.844 | 34.39 | 28.61 | 27.11 |
| PULSr [17] | CVPR/2021 | 6.98 | 4.94 | 5.12 | - | - | - | 32.65 | 34.71 | **34.45** |
| DC-ShadowGAN [3] | CVPR/2022 | 10.55 | 5.79 | 6.57 | 0.976 | 0.958 | 0.922 | 31.69 | 28.99 | 26.38 |
| Zhu et al. [18] | AAAI/2022 | 7.44 | 3.74 | 4.79 | 0.980 | **0.982** | 0.952 | 34.94 | **35.58** | 31.72 |
| BMNet [22] | CVPR/2022 | 7.60 | 4.59 | 5.02 | 0.988 | 0.976 | 0.959 | 35.61 | 32.80 | 30.28 |
| ShadowFormer[8] | CVPR/2023 | 6.16 | 3.90 | 4.27 | - | - | - | - | - | - |
| SADC | 2024 | 7.19 | 5.06 | 5.41 | **0.989** | 0.976 | **0.961** | 35.52 | 31.97 | 29.85 |

| Our Model | - | **6.29** | 5.64 | 3.35 | **0.989** | 0.965 | 0.950 | **35.92** | 30.04 | 28.57 |

The performance of our method was assessed using three widely adopted metrics: RMSE, PSNR, and SSIM. The experiments were carried out on a computing platform equipped with an NVIDIA V100 GPU (32GB), utilizing CUDA 11.8 for acceleration. Python 3.8 was used for implementation, with PyTorch 2.0.0 serving as the deep learning framework.

4.1 Comparison to the State-of-the-Art

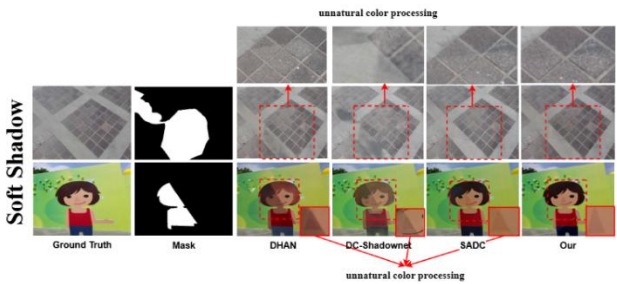

**FIGURE 3.** Comparison of shadow removal results on soft shadow images

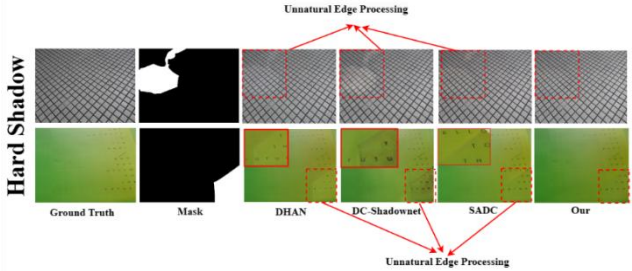

**FIGURE 4.** Comparison of shadow removal from hard shadow images

Our model has been evaluated against state-of-the-art (SOTA) methods from the past three years, particularly those utilizing Generative Adversarial Networks (GANs), on the ISTD, ISTD+, and SRD datasets. The quantitative analysis is detailed in Table 1, where bold values highlight key instances of performance improvement and unnatural color processing observed in other methods, underscoring the advantages of our approach.

As shown in Table 1, our model consistently outperforms existing SOTA methods on the ISTD dataset, achieving the lowest RMSE values of 6.29, 5.64, and 3.35 in the shadow, non-shadow, and overall regions, respectively, significantly reducing shadow residual errors and demonstrating outstanding image reconstruction capabilities. Additionally, in shadow regions, our model achieves an SSIM of 0.989 and a PSNR of 35.92. The high SSIM reflects the model's ability to preserve image textures and edge details, making the de-shadowed image closer to the real shadow-free image in brightness, contrast, and structural detail, particularly achieving natural light-shadow transitions in soft shadow areas. Meanwhile, the improved PSNR indicates the model's effectiveness in reducing artifacts and noise, which is attributed to the design of chromaticity loss and edge loss functions that ensure consistent color restoration and smooth boundary transitions, further enhancing the overall image quality.

Visual comparisons, as shown in Figures 3 and 4, highlight the superiority of our approach in handling both soft and hard shadows. In Figure 3, the red solid-line boxes, which magnify portions of soft shadow regions, demonstrate that our model achieves smoother and more natural color transitions compared to other methods, which often result in abrupt or unnatural shifts. In Figure 4, the red dashed boxes in hard shadow regions reveal that our model produces more seamless edge blending and preserves texture details, while other methods exhibit noticeable artifacts or unnatural edges. These results reflect the effectiveness of our chromaticity and edge loss functions in ensuring consistent color restoration and precise boundary transitions.

4.3 Ablation Study

**TABLE 2.** The quantitative performance results of the single-path model trained only on hard shadows across three datasets

| Dataset | RMSE | | | SSIM | | | PSNR | | |
|---|---|---|---|---|---|---|---|---|---|
| | Shadow | No Shadow | Avg | Shadow | No Shadow | Avg | Shadow | No Shadow | Avg |
| Hard | 7.04 | 7.44 | 4.31 | 0.99 | 0.97 | 0.94 | 34.77 | 27.77 | 26.56 |
| Hard+Soft | 8.81 | 9.41 | 5.47 | 0.99 | 0.96 | 0.94 | 32.66 | 26.01 | 24.60 |
| Soft | 16.51 | 13.19 | 8.30 | 0.94 | 0.88 | 0.82 | 21.65 | 21.65 | 18.15 |

**TABLE 3.** The quantitative performance results of the single-path model trained only on soft shadows across three datasets

| Dataset | RMSE | | | SSIM | | | PSNR | | |
|---|---|---|---|---|---|---|---|---|---|
| | Shadow | No Shadow | Avg | Shadow | No Shadow | Avg | Shadow | No Shadow | Avg |
| Hard | 14.26 | 6.23 | 4.68 | 0.99 | 0.97 | 0.95 | 29.33 | 28.87 | 25.01 |
| Hard+Soft | 14.11 | 9.72 | 6.21 | 0.99 | 0.97 | 0.94 | 29.59 | 27.67 | 24.54 |
| Soft | 19.77 | 23.29 | 12.80 | 0.97 | 0.90 | 0.84 | 25.57 | 22.95 | 20.49 |

**TABLE 4.** The quantitative performance results of the dual-path model trained on both soft and hard shadow datasets across three datasets

| Dataset | RMSE | | | SSIM | | | PSNR | | |
|---|---|---|---|---|---|---|---|---|---|
| | Shadow | No Shadow | Avg | Shadow | No Shadow | Avg | Shadow | No Shadow | Avg |
| Hard | 6.29 | 5.63 | 3.34 | 0.99 | 0.97 | 0.96 | 35.92 | 30.04 | 25.01 |
| Hard+Soft | 8.16 | 7.39 | 4.44 | 0.99 | 0.96 | 0.95 | 33.54 | 28.27 | 24.54 |
| Soft | 15.85 | 12.26 | 7.80 | 0.95 | 0.88 | 0.82 | 22.03 | 22.41 | 20.49 |

Section 2 discusses the differences between soft and hard shadows and the challenges they pose for shadow removal. Soft shadows have diffuse transitions, while hard shadows have sharp, well-defined edges. To address these differences, we developed a dual-path model that processes soft and hard shadows separately. Each path is specifically optimized to handle the unique characteristics of its shadow type, resulting in more accurate shadow removal.

The dual-path architecture uses separate training for each path. The soft shadow path is trained with consistency, adversarial, perceptual, and chromaticity loss functions on a soft shadow dataset. The hard shadow path is trained with consistency, adversarial, perceptual, and shadow edge loss functions on a hard shadow dataset. During training, one path is frozen while the other is optimized to ensure focused learning for each shadow type. Ablation experiments demonstrate that this dual-path design improves performance by addressing the distinct features of soft and hard shadows.

The generator model consists of two paths: one for hard shadows and one for soft shadows. Each path includes a Swin Transformer for feature extraction and a UNet++ for multi-layer feature fusion. The soft shadow path is trained with specific loss functions on a soft shadow dataset, while the hard shadow path is trained with its own set of loss functions on a hard shadow dataset.

To evaluate this design, we conducted several experiments. The control group involved a four-step training process. First, the dual-path model was trained on a mixed dataset using general loss functions. Then, the hard shadow path was fine-tuned using a hard shadow dataset, while the soft shadow path was frozen. Similarly, the soft shadow path was fine-tuned on a soft shadow dataset, with the hard shadow path frozen. Finally, the fusion layer was trained on the mixed dataset, while all other layers were frozen.

For comparison, two experimental groups were used. The first trained a single-path model on a hard shadow dataset, and the second trained a single-path model on a soft shadow dataset. All models were trained for the same number of epochs and tested on mixed, hard shadow, and soft shadow datasets.

To support training and evaluation, two separate datasets were created. The hard shadow dataset, sourced from the ISTD dataset, includes images with sharp, well-defined shadow edges. The soft shadow dataset, derived from the expanded LRSS dataset, contains images with diffuse, less-defined shadows. A mixed dataset was also created by combining images from both the ISTD and LRSS datasets to ensure balanced representation and comprehensive evaluation.

The dual-path model was evaluated on three datasets: the hard shadow dataset, the soft shadow dataset, and the mixed dataset. As shown in Tables 2, 3, and 4, the dual-path model outperformed the single-path models across all datasets.

The dual-path model achieved lower root mean square error (RMSE) values in shadow, non-shadow, and overall regions compared to single-path models, indicating its shadow-free images were closer to the ground truth. It also achieved higher peak signal-to-noise ratio (PSNR) values, showing better accuracy in image restoration. While differences in structural similarity index measure (SSIM) were small, the dual-path model consistently scored higher, reflecting better preservation of brightness, contrast, and structural details.

Additionally, the dual-path model trained on the mixed dataset showed strong generalization, performing better on both hard and soft shadow datasets compared to single-path models trained on only one type of shadow. This highlights a limitation of single-path models, which struggle to handle mixed shadow characteristics.

Shadows in real-world images often blend features of both soft and hard shadows, with no clear boundary between them. By integrating data from both shadow types and using tailored loss functions, the dual-path model effectively learns to handle these mixed characteristics. This approach not only improves flexibility and accuracy but also enhances the model's practical applicability in complex environments.

## 5. Conclusion

This study proposed a novel shadow removal method that processes hard and soft shadows separately using a dual-path model. The model incorporates a classifier to determine the shadow type and then removes it through specialized pathways designed for each type. The hard shadow path targets shadows with sharp edges, leveraging an edge loss function, while the soft shadow path addresses blurred-edge shadows using a chromaticity loss function. Our method demonstrated superior performance compared to existing state-of-the-art models on the ISTD dataset, establishing a new benchmark. Future work will focus on evaluating and refining the model on additional datasets to improve its generalization and applicability. This research not only advances shadow removal techniques but also provides valuable insights for broader computer vision tasks.

## References

[1] B. Ding, C. Long, and C. Xiao, "Argan: Attentive recurrent generative adversarial network for shadow detection and removal," in Proc. of the IEEE/CVF International Conference on Computer Vision (ICCV), 2019.

[2] O. Sidorov, "Conditional GANs for Multi-Illuminant Color Constancy: Revolution or Yet Another

Approach?" in Proc. IEEE Conf. on Computer Vision and Pattern Recognition (CVPR).

[3] Y. Jin, A. Sharma, and R. T. Tan, "DC-ShadowNet: Single-Image Hard and Soft Shadow Removal Using Unsupervised Domain-Classifier Guided Network," in Proc. of the IEEE/CVF International Conference on Computer Vision, pp. 5027–5036, 2021.

[4] L. Qu, J. Tian, S. He, Y. Tang, and R. W. H. Lau, "Deshadownet: A multi-context embedding deep network for shadow removal," in Proc. of the IEEE Conference on Computer Vision and Pattern Recognition (CVPR), 2017.

[5] L. Qu, J. Tian, S. He, Y. Tang, and R. W. H. Lau, "DeshadowNet: A Multi-context Embedding Deep Network for Shadow Removal," in Proc. IEEE Conf. on Computer Vision and Pattern Recognition (CVPR), 2017.

[6] J. Wang, X. Li, and J. Yang, "Stacked conditional generative adversarial networks for jointly learning shadow detection and shadow removal," in Proc. of the IEEE Conference on Computer Vision and Pattern Recognition (CVPR), 2018, pp. 1788–1797.

[7] X. Hu, Y. Jiang, C.-W. Fu, and P.-A. Heng, "Mask-shadowgan: Learning to remove shadows from unpaired data," in Proc. of the IEEE/CVF International Conference on Computer Vision (ICCV), 2019, pp. 2472–2481.

[8] N. K. Logothetis and D. L. Sheinberg, "Visual Object Recognition," in Annual Review of Neuroscience, vol. 19, pp. 577-621, 1996.

[9] J. Wang, X. Li, L. Hui, and J. Yang, "Stacked Conditional Generative Adversarial Networks for 20Jointly Learning Shadow Detection and Shadow Removal," in Proc. IEEE/CVF Conf. Comput. Vis. Pattern Recognit. (CVPR), 2021.

[10] R. Abiko and M. Ikehara, "Channel Attention GAN Trained With Enhanced Dataset for Single-Image Shadow Removal," IEEE Access, vol. 10, 2022, doi: 10.1109/ACCESS.2022.3147063.

[11] L. Zhang, C. Long, X. Zhang, and C. Xiao, "RIS-GAN: Explore Residual and Illumination with Generative Adversarial Networks for Shadow Removal," in Proc. of the Thirty-Fourth AAAI Conference on Artificial Intelligence (AAAI-20), Wuhan University of Science and Technology; Kitware Inc.; School of Computer Science, Wuhan University, 2020.

[12] X. Cun, C.-M. Pun, and C. Shi, "Towards Ghost-Free Shadow Removal via Dual Hierarchical Aggregation Network and Shadow Matting GAN," in Proc. of the Thirty-Fourth AAAI Conference on Artificial Intelligence (AAAI-20), University of Macau; Xi'an University of Technology, 2020.

[13] J. Wang, L. Hui, X. Li, and J. Yang, "Stacked Conditional Generative Adversarial Networks for Jointly Learning Shadow Detection and Shadow Removal," arXiv preprint arXiv:1712.02478, 2017.

[14] Z. Yang, S. Ren, Z. Wu, N. Zhao, J. Wang, J. Qin, and S. He, "NPF-200: A Multi-Modal Eye Fixation Dataset and Method for Non-Photorealistic Videos," in *Proc. ACM Int. Conf. Multimedia (ACM MM), 2023.42

[15] M. Gryka, M. Terry, and G. J. Brostow, "Learning to Remove Soft Shadows," in ACM Transactions on Graphics, vol. XX, no. X, pp. XX-XX, 2015.

[16] X. Cun, C.-M. Pun, and C. Shi, "Towards ghost-free shadow removal via dual hierarchical aggregation network and shadow matting gan," in Proc. of the AAAI Conference on Artificial Intelligence, vol. 34, no. 07, pp. 10,680–10,687, 2020.

[17] F.-A. Vasluianu, A. Romero, L. Van Gool, and R. Timofte, "Shadow removal with paired and unpaired learning," in Proc. of the IEEE/CVF Conference on Computer Vision and Pattern Recognition, pp. 826–835, 2021.

[18] Y. Zhu, Z. Xiao, Y. Fang, X. Fu, Z. Xiong, and Z.-J. Zha, "Efficient Model-Driven Network for Shadow Removal," 2022.

[19] L. Fu, C. Zhou, Q. Guo, F. Juefei-Xu, H. Yu, W. Feng, Y. Liu, and S. Wang, "Auto-exposure fusion for single-image shadow removal," in Proc. of the IEEE/CVF International Conference on Computer Vision, pp. 10,571–10,580, 2021.

[20] L. Zhang, C. Long, X. Zhang, and C. Xiao, "Ris-gan: Explore residual and illumination with generative adversarial networks for shadow removal," in Proc. of the AAAI Conference on Artificial Intelligence, vol. 34, no. 07, pp. 12,829–12,836, 2020.

[21] L. Zhang, C. Long, Q. Yan, X. Zhang, and C. Xiao, "Cla-gan: A context and lightness aware generative adversarial network for shadow removal," in Computer Graphics Forum, vol. 39, no. 7, pp. 483–494, Wiley Online Library, 2020.

[22] Y. He, Y. Xing, T. Zhang, and Q. Chen, "Bijective mapping network for shadow removal," in Proc. of the IEEE/CVF Conference on Computer Vision and Pattern Recognition, pp. 5627–5636, 2022.

[23] Y. Xu, M. Lin, H. Yang, K. Li, Y. Shen, F. Chao, and R. Ji, "Shadow-aware dynamic convolution for shadow removal," Pattern Recognition, vol. 112, no. 3, pp. 107-115, 2022.

[24] X. Hu, Y. Jiang, C.-W. Fu, and P.-A. Heng, "Mask-ShadowGAN: Learning to Remove Shadows from Unpaired Data," in Proc. of the IEEE/CVF International Conference on Computer Vision (ICCV), 2019. Available: https://arxiv.org/abs/1903.10683

[25] R. Guo, Q. Dai, and D. Hoiem, "Paired regions for shadow detection and removal," IEEE Transactions on Pattern Analysis and Machine Intelligence, vol. 35, no. 12, pp. 2956–2967, 2012.